\documentclass{article}
\usepackage{inputenc}
\usepackage{tcolorbox}
\usepackage{pgfkeys}
\usepackage{hyperref}
\usepackage{amsmath, amssymb, amsthm}
\usepackage{microtype, graphicx, subfigure, booktabs}
\usepackage{xcolor, listings}
\usepackage{enumitem}
\usepackage{mdframed}
\usepackage{multirow, adjustbox, wrapfig, array}
\tcbuselibrary{skins}

\usepackage[accepted]{icml2024}

\usepackage{amsmath}
\usepackage{amssymb}
\usepackage{amsthm}

\usepackage{xcolor}
\usepackage{courier}
\usepackage{listings}
\usepackage{tcolorbox}

\usepackage{booktabs}
\usepackage{multirow}
\usepackage{adjustbox}
\usepackage{wrapfig}
\usepackage{array}
\usepackage{multirow}

\definecolor{bgcolor}{rgb}{0.95, 0.95, 0.95}
\definecolor{codegray}{rgb}{0.5, 0.5, 0.5}
\definecolor{codepurple}{rgb}{0.58, 0.44, 0.86}
\definecolor{codegreen}{rgb}{0.44, 0.74, 0.47}
\definecolor{codeblue}{rgb}{0.35, 0.6, 0.8}
\definecolor{codeturquoise}{rgb}{0.3, 0.8, 0.8}
\definecolor{codetext}{rgb}{0.1, 0.1, 0.1}
\definecolor{framecolor}{rgb}{0.8, 0.8, 0.8}

\newtcolorbox{shadowbox}[2][]{
  colback=white,
  colframe=black!70,
  boxrule=0.5pt,
  sharp corners,
  title=#2,
  fonttitle=\bfseries,
  #1
}

\usepackage{mdframed}

\newcounter{question}
\usepackage{mdframed}

\newmdenv[
  leftmargin=1cm,
  rightmargin=1cm,
  skipabove=\topsep,
  skipbelow=\topsep
]{factorbox}

\icmltitlerunning{Open Artificial Knowledge Dataset}

\begin{document}

\twocolumn[
\icmltitle{Open Artificial Knowledge}




\begin{icmlauthorlist}
\icmlauthor{Vadim Borisov}{ttt}
\icmlauthor{Richard H. Schreiber}{ttt}
\end{icmlauthorlist}

\icmlaffiliation{ttt}{\href{http://tabularis.ai}{\textbf{tabularis.ai}}}
\icmlcorrespondingauthor{Vadim Borisov}{vadim@tabularis.ai}

\icmlkeywords{Machine Learning, ICML}

\vskip 0.3in
]



\printAffiliationsAndNotice{\icmlEqualContribution} 

\newcommand\sbullet[1][.5]{\mathbin{\vcenter{\hbox{\scalebox{#1}{$\bullet$}}}}}

\begin{abstract}

The tremendous success of chat-based AI systems like ChatGPT, Claude, and Gemini stems from Large Language Models (LLMs) trained on vast amount of datasets. However, acquiring high-quality, diverse, and ethically sourced training data remains a significant challenge. We introduce the \textbf{O}pen \textbf{A}rtificial \textbf{K}nowledge (\textbf{OAK}) dataset, a large-scale resource of \textbf{over 500 million tokens}  \textit{(at the moment of writing)} designed to address this issue. OAK leverages an ensemble of state-of-the-art LLMs, including GPT4o, LLaMa3-70B, LLaMa3-8B, Mixtral-8x7B, Gemma-7B, and Gemma-2-9B , to generate high-quality text across diverse domains, guided by Wikipedia's main categories. Our methodology ensures broad knowledge coverage while maintaining coherence and factual accuracy. The OAK dataset aims to foster the development of more capable and aligned language models while addressing critical issues of data scarcity and privacy in LLM training, and it is freely available on \href{http://oakdataset.org/}{\textbf{oakdataset.org}}.

\end{abstract}

\begin{figure*}[ht]
    \centering
    \includegraphics[trim=5.0cm 13.2cm 5.0cm 12.0cm, clip, width=\textwidth]{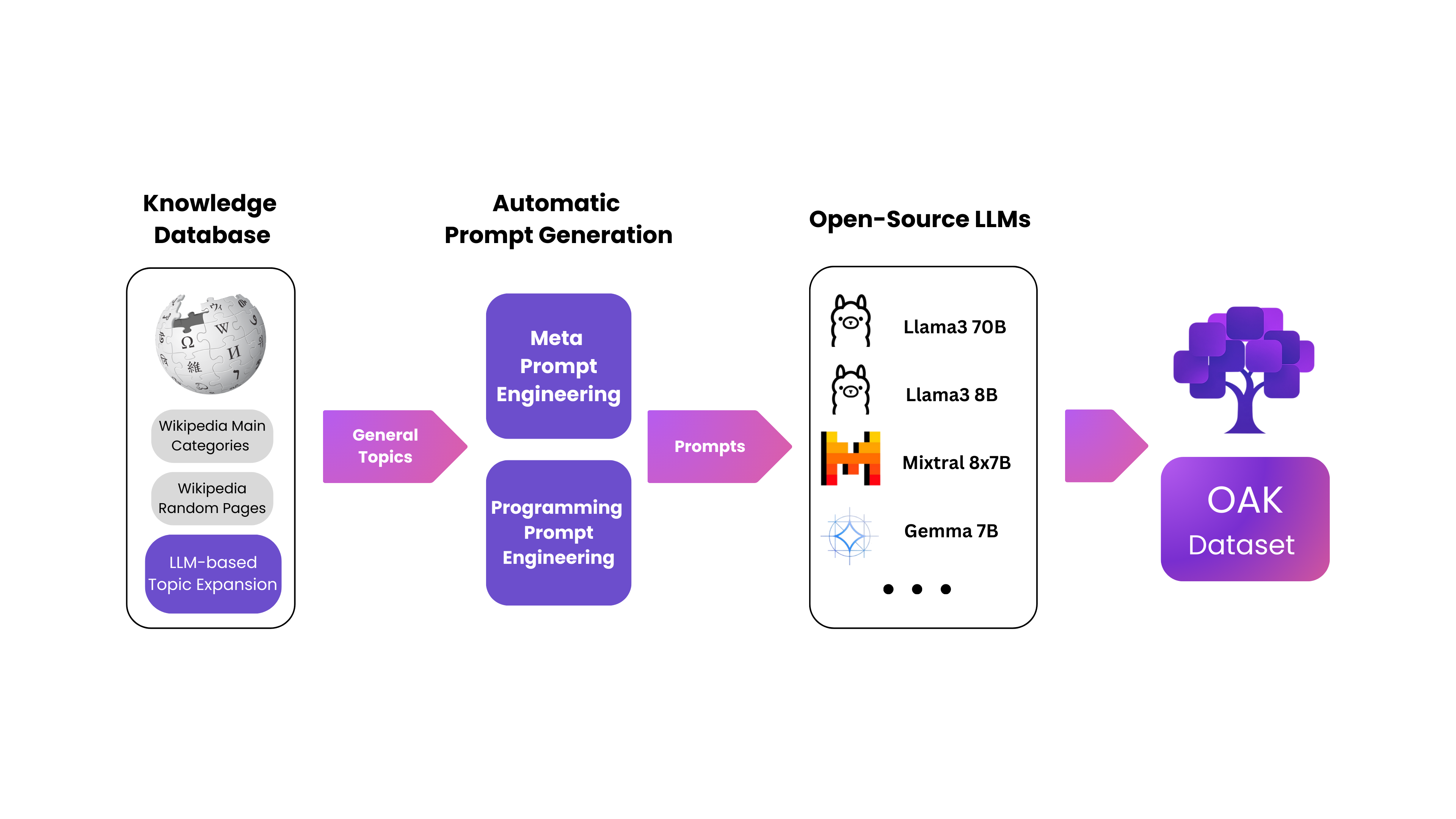}
    \caption{Overview of the Open Artificial Knowledge (OAK) dataset generation pipeline. The process begins with extracting general topics from extensive human knowledge databases such as Wikipedia and GPT-4o models. These high-level and sub-level topics are then used in an automatic prompt generation step, which employs two methods: meta prompt engineering using large language models (LLMs) and cost-effective programming prompt engineering. The generated prompts are subsequently fed into state-of-the-art open-source LLMs (at the time of writing, five models were used: Llama3-8B, Llama-70B, Mixtral7x8B, Gemma-7B \cite{team2024gemma}, and Gemma-2-9B \cite{gemma_2024}) to create the OAK dataset.}
    \label{fig:pipeline}
\end{figure*}


\section{Introduction}
\label{intro}
The rapid advancement in Artificial Intelligence (AI) and Machine Learning (ML) has underscored the critical need for large, diverse, and high-quality datasets to train and evaluate foundation models \cite{bommasani2021opportunities}. However, acquiring such datasets presents significant challenges, including data scarcity, privacy concerns, and high costs associated with data collection and annotation.
Artificial (synthetic) data has emerged as a promising solution to these challenges, offering a way to generate data that mimics real-world patterns and characteristics \cite{benallal2024cosmopedia, liu2024best, sun2023moss, li2023synthetic, long2024llms, borisov2022language}. The importance of artificial data in AI research has grown substantially due to several factors:
\begin{itemize}
    \item[--] \textit{Scalability:} Synthetic data can be generated at scale, addressing the need for massive datasets required by modern AI models.
    \item[--] \textit{Privacy preservation:} Artificial data can help mitigate privacy issues by creating anonymized datasets free from sensitive personal information.
    \item[--] \textit{Diversity and representation:} Synthetic data can be controlled (conditioned) to cover a wide range of scenarios, potentially addressing biases present in real-world datasets.
    \item[--] \textit{Cost-effectiveness:} Generating artificial data can be more economical than collecting and annotating real-world data.
\end{itemize}





The use of synthetic datasets in training state-of-the-art language models (LLMs) has become increasingly prevalent. This trend is evident in models like Llama-3\footnote{\url{https://llama.meta.com/llama3/}}, which built upon its predecessor, Llama 2 \cite{touvron2023llama}, by incorporating synthetic data in its training process. Similar approaches have been applied to other advanced models \cite{young2024yi, li2023textbooks, tunstall2023zephyr}.

While handcrafted human data has shown significant improvements in \textit{supervised fine-tuning} (SFT) of LLMs, particularly for tasks like code generation and mathematical reasoning \cite{roziere2023code, wan2024deep}, the scarcity and cost of creating such high-quality data have led to the increasing use of synthetic data as a proxy. This method primarily leverages strongly capable LLMs, such as the GPT family \cite{achiam2023gpt} to produce high-quality synthetic data \cite{li2023synthetic, josifoski2023exploiting, taori2023stanford}.

Recent research has highlighted LLMs' ability to rephrase for improved responses and boost synthetic data for effective SFT \cite{gallego2024refined, chen2024self}. These developments suggest that the use of synthetic data in model training will continue to grow in the future, with ongoing research exploring various techniques to leverage synthetic data effectively for improving LLM performance and alignment \cite{hao2024synthetic}.



%



\section{Key Challenges of Artificial Data}
\label{sec:challenges}

The generation of artificial data presents key challenges and considerations essential for its efficacy and ethical use. These include diversity, quality, privacy, bias, and broader ethical and legal issues. Additionally, practical tasks such as scalability, evaluation metrics, cost-effectiveness, integration with real data, and factual accuracy are critical for the effective use of synthetic data. Below we describe the main challenges in details:
\vspace{-0.25cm}
\begin{itemize}[label=$\bigstar$, ]
\item \textbf{C1} \textit{Diversity and Generalization} 
Ensuring sufficient diversity in artificial data is crucial to enable model generalization. Artificial data must encompass a wide range of scenarios and variations to prevent models from overfitting to specific patterns or biases inherent in the synthetic dataset \cite{long2024llms}. For example, an LLM trained only on synthetic texts that lack cultural or linguistic variety may fail to understand and generate responses for diverse real-world inputs.

\item \textbf{C2} \textit{Quality} 
The quality of synthetic data directly impacts the performance of the models trained on it. High-quality synthetic data should closely mimic the characteristics of real-world data, maintaining consistency, relevance, and richness in the generated information. For instance, synthetic conversational data should exhibit natural dialogue flow and contextually appropriate responses.

\item \textbf{C3} \textit{Privacy} 
Artificial data generation offers a solution to privacy concerns by reducing the dependency on real, potentially sensitive data. However, it is crucial to ensure that the synthetic data itself does not inadvertently reveal any sensitive information or patterns that could compromise privacy. 

\item \textbf{C4} \textit{Bias} 
Bias in artificial data can arise from the underlying algorithms and training data used to generate it. Addressing bias is essential to avoid perpetuating or amplifying existing biases in the real-world data, which can lead to unfair or inaccurate model predictions. For example, synthetic text data should be carefully monitored to avoid reinforcing gender or racial stereotypes \cite{hao2024synthetic}.

\item \textbf{C5} \textit{Ethical and Legal Considerations} 
The creation and use of synthetic data must adhere to ethical guidelines and legal regulations. This includes ensuring transparency in the data generation process, obtaining necessary permissions, and avoiding the misuse of synthetic data in ways that could harm individuals or society. Regulatory frameworks such as General Data Protection Regulation (GDPR)~\cite{GDPR} or California Consumer Privacy Act (CCPA) \cite{ccpa2021} provide guidelines for the ethical and legal use of synthetic textual data.

\item \textbf{C6} \textit{Toxicity and Harmful Content} 
Artificial data must be free from toxic or harmful content to ensure the safety and well-being of users \cite{li2023synthetic}. This involves rigorous filtering and monitoring processes to detect and eliminate offensive, inappropriate, or harmful text. For instance, synthetic conversational data should be carefully screened to prevent the generation of abusive or discriminatory language.

\end{itemize}

Furthermore, several practical tasks are critical to the successful generation and application of artificial data:

\begin{itemize}[label=$\bigstar$]

\item \textbf{C7} \textit{Scalability and Cost-Effectiveness} 
Producing synthetic data at a scale sufficient to train large models effectively, while maintaining quality, is a significant challenge \cite{long2024llms}. Efficient data generation techniques must be employed to balance quantity with quality and to ensure cost-effectiveness. This is particularly important for training Large Language Models (LLMs), which require vast and diverse text corpora. While generating high-quality synthetic text data can be resource-intensive, it can reduce the need for expensive data collection and annotation processes.

\item \textbf{C8} \textit{Evaluation Metrics} 
Developing robust metrics and methods for assessing the quality and effectiveness of synthetic data is essential \cite{feng2023factkb, long2024llms}. These metrics should evaluate how well the synthetic data supports model training and performance, ensuring it meets the desired standards. 

\item \textbf{C9} \textit{Factual Accuracy} 
Ensuring that synthetic data accurately reflects real-world information without introducing inaccuracies or hallucinations is vital \cite{wei2024long}. The generated data should be as factual and reliable as possible to maintain the integrity of the models trained on it. For instance, synthetic news articles should contain accurate and up-to-date information to avoid misinformation.

\item \textbf{C10} \textit{Maintenance and Update of Synthetic Data} 
As real-world scenarios and language use evolve, synthetic data must be continually updated to remain relevant and useful. This includes addressing the ongoing need to generate new synthetic data that reflects current trends, topics, and linguistic changes. Failure to regularly update synthetic datasets can lead to models that are outdated and less effective. For example, a language model trained on outdated synthetic text data may not understand or generate responses about new technologies, cultural shifts, or emerging terminologies.

\end{itemize}

\section{OAK Dataset}
\label{sec:oak_dataset}

The Open Artificial Knowledge (OAK) dataset generation follows a structured approach (Fig. \ref{fig:pipeline}) designed to address the key challenges of artificial data creation as outlined in Section \ref{sec:challenges}. Below we list of the main steps for the OAK dataset creation:

\begin{enumerate}[label=\textbf{Step \arabic*:}, leftmargin=3.5em]
    \item \textit{\textbf{Subject Extraction}}
    Using human knowledge databases like Wikipedia \cite{wikipedia}, we extract high-level topics. This step directly addresses \textit{Diversity and Generalization} (\textbf{C1}) by covering a broad range of categories to prevent model overfitting and enhance generalization capabilities.

    \item \textit{\textbf{Subtopic Expansion}}
    These high-level topics are expanded with subtopics through advanced language models such as OpenAI's GPT-4o~\cite{achiam2023gpt}, improving further the \textbf{C1} and \textit{Quality} (\textbf{C2}) by adding depth and breadth that mimic real-world data variability. In total, we generated $493,237$ unique subcategories given $21,311$ categories extracted from Wikipedia \cite{wikipedia}. A detailed explanation of the subtopic generation is in Appendix \ref{app:subtopic}.  

    \item \textit{\textbf{Prompt Generation}} 
    Prompts are generated employing the programming prompt engineering and meta-prompt technique, which are tuned to optimize the prompts' quality, length, and style, also addressing \textbf{C1}, \textbf{C2} challenges. This step also tackles \textit{Bias} (\textbf{C4}) by carefully generating prompts with different conditions to mitigate the introduction of bias. Furthermore, by utilizing highly aligned LLMs models we can deals with \textit{Factual Accuracy} (\textbf{C9}).  

    \item \textit{\textbf{Text Generation with Open-Source LLMs}}
    To tackle the diversity challenge \textbf{C1} even further, we utilize several open-source LLMs, like Llama3-8b, Llama-70b, Mitral7x8b, Gemma-7b, and Gemma2-9B. This phase tackles also \textit{Scalability and Cost-Effectiveness} (\textbf{C7}) by using efficient, open-source models to produce large volumes of data. Examples of the generated text are in Appendix (Figs. \ref{fig:sample_1}, \ref{fig:sample_2}, \ref{fig:sample_3}).

\end{enumerate}

To address the \textit{Privacy} (\textbf{C3}) challenge, we implement a multi-faceted approach. We exclusively use publicly available data and open-source LLM models, ensuring the dataset is free from private content. 
For the \textit{Ethical and Legal Considerations} (\textbf{C5}) challenge, we have implemented a comprehensive strategy. All code is published online, promoting transparency and reproducibility. We are committed to promptly removing any content upon request, ensuring compliance with ethical guidelines and individual rights. We continuously monitor the dataset's applications in the research community to identify and address any emerging ethical concerns.

Addressing the \textit{Toxicity and Harmful Content} (\textbf{C6}) challenge involves using automated filtering techniques, such as basic natural language processing methods to filter inappropriate content (e.g., regex for personal identifiers). Furthermore, we fine-tune the ELECTRA model \cite{clark2020electra} on publicly available toxicity datasets \cite{adams2017toxic} to provide a filtering score.

For the \textit{Evaluation Metrics} (\textbf{C8}) challenge, we will engage the community to fine-tune an LLM on the OAK dataset and evaluate it using common benchmarks such as WinoGrande \cite{sakaguchi2021winogrande}, ARC Easy \cite{clark2018arc}, and so forth, the list of the benchmarks will be extended.

We plan regular updates of the OAK dataset to reflect new trends and information, addressing the \textit{Maintenance and Update of Synthetic Data} (\textbf{C10}). This step ensures the dataset remains relevant and effective for training purposes.

The data generation pipeline is illustrated in Fig. \ref{fig:pipeline}, starting with the querying of knowledge databases to gather topics, which are then expanded using LLMs. These topics are transformed into prompts that are subsequently used to generate texts with state-of-the-art models. The resulting OAK dataset is continuously evaluated and updated, ensuring its effectiveness and reliability for training advanced language models. By systematically addressing each challenge, the OAK dataset provides a robust resource for developing more accurate and aligned language models.


\section{Automatic Prompt Generation}
\label{sec:prompt_engineering}

In this section, we discuss the main techniques we used to generate the OAK dataset. One of the most challenging aspects of working with LLMs is prompt tuning \cite{schulhoff2024prompt}; since, prompts play a crucial role in determining how well the model performs. We utilized zero-shot and few-shot techniques like chain-of-thoughts (CoT) \cite{wei2022chain} and emotional prompts \cite{li2023large} to enhance effectiveness.

\subsection{Programming Prompt Engineering}
To generate programmable prompts, we employ a versatile template-based approach that allows for systematic variation of key parameters \cite{prompt_engineering2023}. By randomly altering the length, style, and type of analysis within these templates, we create a diverse set of prompts that can effectively train and test language models across a wide range of scenarios \cite{prompt_design2023}. Figure \ref{fig:programming_prompt} illustrates a pseudocode example of our programming prompt engineering process. This approach ensures that our synthetic data encompasses the necessary variability to enhance model robustness and generalization. However, a primary limitation of this approach is that the resulting generated prompts may lack realism.

\subsection{Meta Prompt Engineering}

To overcome the limitations of realism in programming prompt engineering, we employ the meta prompt engineering technique, which uses advanced LLMs to generate and refine prompts conditional on quality, length, and style \cite{reynolds2021prompt, zhou2022large, ye2023prompt}. We utlize few-shot learning and chain-of-thought (CoT) techniques \cite{wei2022chain} to guide the LLMs in creating appropriate prompts. The process incorporates emotional prompting \cite{li2023large} to enhance the richness and variety of the generated content. By continuously refining and evaluating the generated prompts, we ensure they meet specific quality criteria and address potential biases.

This method allows for the creation of a vast array of high-quality, contextually appropriate prompts that cover a wide range of topics and styles. The meta prompt engineering approach addresses key challenges such as generalization (\textbf{C1}), quality (\textbf{C2}), and bias mitigation (\textbf{C4}) in artificial data creation, enhancing the OAK dataset's utility for training robust language models. Figure \ref{fig:meta_prompt} displays our selected meta prompt, which serves as the foundation for generating the diverse array of prompts used in creating the OAK dataset.

\section{Use Considerations}
We release the Open Artificial Knowledge (OAK) dataset to accelerate open LLM research, particularly in areas such as model alignment \cite{shen2023large}, bias mitigation, prompt engineering \cite{prompt_design2023}, and so forth. The OAK dataset and associated tools are intended for research purposes only. Researchers and practitioners are encouraged to use OAK for developing and fine-tuning language models, evaluating model performance, and exploring advanced NLP applications. However, users must adhere to ethical guidelines, respect privacy considerations, and be mindful of potential biases in the synthetic data. The authors are committed to regularly updating the dataset and removing any content upon request to maintain its relevance and ethical standards.

\section{Conclusion and Future Work}

We present the Open Artificial Knowledge (OAK) dataset, a comprehensive resource for AI research derived from Wikipedia's main categories. OAK leverages advanced models like GPT4o, LLaMa3, Mixtral, Gemma, and Gemma2 to address data scarcity, privacy, and diversity issues. With over 500 million tokens, this freely available dataset supports model alignment, fine-tuning, and benchmarking across a wide range of AI tasks and applications.

Future work will focus on expanding linguistic diversity and accessibility, incorporating advanced models for data generation, and refining code-related tasks by integrating code-centric datasets. We aim to develop a framework for community contributions. These efforts will further enrich OAK, enhancing its utility across various AI applications and research domains, while continuously adapting to emerging trends and challenges in the field.

\nocite{langley00}

\bibliography{MAIN}

\begin{thebibliography}{39}
\providecommand{\natexlab}[1]{#1}
\providecommand{\url}[1]{\texttt{#1}}
\expandafter\ifx\csname urlstyle\endcsname\relax
  \providecommand{\doi}[1]{doi: #1}\else
  \providecommand{\doi}{doi: \begingroup \urlstyle{rm}\Url}\fi

\bibitem[Achiam et~al.(2023)Achiam, Adler, Agarwal, Ahmad, Akkaya, Aleman, Almeida, Altenschmidt, Altman, Anadkat, et~al.]{achiam2023gpt}
Achiam, J., Adler, S., Agarwal, S., Ahmad, L., Akkaya, I., Aleman, F.~L., Almeida, D., Altenschmidt, J., Altman, S., Anadkat, S., et~al.
\newblock Gpt-4 technical report.
\newblock \emph{arXiv preprint arXiv:2303.08774}, 2023.

\bibitem[Ben~Allal et~al.(2024)Ben~Allal, Lozhkov, Penedo, Wolf, and von Werra]{benallal2024cosmopedia}
Ben~Allal, L., Lozhkov, A., Penedo, G., Wolf, T., and von Werra, L.
\newblock Cosmopedia, feb 2024.
\newblock URL \url{https://huggingface.co/datasets/HuggingFaceTB/cosmopedia}.

\bibitem[Bommasani et~al.(2021)Bommasani, Hudson, Adeli, Altman, Arora, von Arx, Bernstein, Bohg, Bosselut, Brunskill, et~al.]{bommasani2021opportunities}
Bommasani, R., Hudson, D.~A., Adeli, E., Altman, R., Arora, S., von Arx, S., Bernstein, M.~S., Bohg, J., Bosselut, A., Brunskill, E., et~al.
\newblock On the opportunities and risks of foundation models.
\newblock \emph{arXiv preprint arXiv:2108.07258}, 2021.

\bibitem[Borisov et~al.(2022)Borisov, Se{\ss}ler, Leemann, Pawelczyk, and Kasneci]{borisov2022language}
Borisov, V., Se{\ss}ler, K., Leemann, T., Pawelczyk, M., and Kasneci, G.
\newblock Language models are realistic tabular data generators.
\newblock \emph{arXiv preprint arXiv:2210.06280}, 2022.

\bibitem[Chen et~al.(2024)Chen, Deng, Yuan, Ji, and Gu]{chen2024self}
Chen, Z., Deng, Y., Yuan, H., Ji, K., and Gu, Q.
\newblock Self-play fine-tuning converts weak language models to strong language models.
\newblock \emph{arXiv preprint arXiv:2401.01335}, 2024.

\bibitem[Clark et~al.(2020)Clark, Luong, Le, and Manning]{clark2020electra}
Clark, K., Luong, M.-T., Le, Q.~V., and Manning, C.~D.
\newblock Electra: Pre-training text encoders as discriminators rather than generators.
\newblock \emph{arXiv preprint arXiv:2003.10555}, 2020.

\bibitem[Clark et~al.(2018)Clark, Cowhey, Etzioni, Khot, Sabharwal, Schoenick, and Tafjord]{clark2018arc}
Clark, P., Cowhey, I., Etzioni, O., Khot, T., Sabharwal, A., Schoenick, C., and Tafjord, O.
\newblock Think you have solved question answering? try arc, the ai2 reasoning challenge, 2018.
\newblock URL \url{https://arxiv.org/abs/1803.05457}.
\newblock Allen Institute for Artificial Intelligence (AI2).

\bibitem[European~Parliament(2016)]{GDPR}
European~Parliament, C. o. t. E.~U.
\newblock Regulation (eu) 2016/679 of the european parliament and of the council of 27 april 2016.
\newblock \emph{Official Journal of the European Union}, 2016.

\bibitem[Feng et~al.(2023)Feng, Balachandran, Bai, and Tsvetkov]{feng2023factkb}
Feng, S., Balachandran, V., Bai, Y., and Tsvetkov, Y.
\newblock Factkb: Generalizable factuality evaluation using language models enhanced with factual knowledge.
\newblock \emph{arXiv preprint arXiv:2305.08281}, 2023.

\bibitem[Gallego(2024)]{gallego2024refined}
Gallego, V.
\newblock Refined direct preference optimization with synthetic data for behavioral alignment of llms.
\newblock \emph{arXiv preprint arXiv:2402.08005}, 2024.

\bibitem[Hao et~al.(2024)Hao, Han, Jiang, Li, Wu, Zhong, Zhou, and Tang]{hao2024synthetic}
Hao, S., Han, W., Jiang, T., Li, Y., Wu, H., Zhong, C., Zhou, Z., and Tang, H.
\newblock Synthetic data in ai: Challenges, applications, and ethical implications.
\newblock \emph{arXiv preprint arXiv:2401.01629}, 2024.

\bibitem[Jiang(2023)]{prompt_engineering2023}
Jiang, X. e.~a.
\newblock Efficient prompting methods for large language models: A survey.
\newblock \emph{arXiv preprint arXiv:2404.01077}, 2023.

\bibitem[Josifoski et~al.(2023)Josifoski, Sakota, Peyrard, and West]{josifoski2023exploiting}
Josifoski, M., Sakota, M., Peyrard, M., and West, R.
\newblock Exploiting asymmetry for synthetic training data generation: Synthie and the case of information extraction.
\newblock \emph{arXiv preprint arXiv:2303.04132}, 2023.

\bibitem[Langley(2000)]{langley00}
Langley, P.
\newblock Crafting papers on machine learning.
\newblock In Langley, P. (ed.), \emph{Proceedings of the 17th International Conference on Machine Learning (ICML 2000)}, pp.\  1207--1216, Stanford, CA, 2000. Morgan Kaufmann.

\bibitem[Li et~al.(2023{\natexlab{a}})Li, Bubeck, Eldan, Del~Giorno, Gunasekar, and Lee]{li2023textbooks}
Li, Y., Bubeck, S., Eldan, R., Del~Giorno, A., Gunasekar, S., and Lee, Y.~T.
\newblock Textbooks are all you need ii: phi-1.5 technical report.
\newblock \emph{arXiv preprint arXiv:2309.05463}, 2023{\natexlab{a}}.

\bibitem[Li et~al.(2023{\natexlab{b}})Li, Zhu, Lu, and Yin]{li2023synthetic}
Li, Z., Zhu, H., Lu, Z., and Yin, M.
\newblock Synthetic data generation with large language models for text classification: Potential and limitations.
\newblock \emph{arXiv preprint arXiv:2310.07849}, 2023{\natexlab{b}}.

\bibitem[Liu et~al.(2024)Liu, Wei, Liu, Si, Zhang, Rao, Zheng, Peng, Yang, Zhou, et~al.]{liu2024best}
Liu, R., Wei, J., Liu, F., Si, C., Zhang, Y., Rao, J., Zheng, S., Peng, D., Yang, D., Zhou, D., et~al.
\newblock Best practices and lessons learned on synthetic data for language models.
\newblock \emph{arXiv preprint arXiv:2404.07503}, 2024.

\bibitem[Long et~al.(2024)Long, Wang, Xiao, Zhao, Ding, Chen, and Wang]{long2024llms}
Long, L., Wang, R., Xiao, R., Zhao, J., Ding, X., Chen, G., and Wang, H.
\newblock On llms-driven synthetic data generation, curation, and evaluation: A survey.
\newblock \emph{arXiv preprint arXiv:2406.15126}, 2024.

\bibitem[Lu et~al.(2023)]{prompt_design2023}
Lu, S. et~al.
\newblock Prompt design and engineering: Introduction and advanced methods.
\newblock \emph{arXiv preprint arXiv:2401.14423}, 2023.

\bibitem[OAG(2021)]{ccpa2021}
OAG, C.
\newblock Ccpa regulations: Final regulation text.
\newblock \emph{Office of the Attorney General, California Department of Justice}, 2021.

\bibitem[Reynolds \& McDonell(2021)Reynolds and McDonell]{reynolds2021prompt}
Reynolds, L. and McDonell, K.
\newblock Prompt programming for large language models: Beyond the few-shot paradigm.
\newblock In \emph{Extended abstracts of the 2021 CHI conference on human factors in computing systems}, pp.\  1--7, 2021.

\bibitem[Roziere et~al.(2023)Roziere, Gehring, Gloeckle, Sootla, Gat, Tan, Adi, Liu, Remez, Rapin, et~al.]{roziere2023code}
Roziere, B., Gehring, J., Gloeckle, F., Sootla, S., Gat, I., Tan, X.~E., Adi, Y., Liu, J., Remez, T., Rapin, J., et~al.
\newblock Code llama: Open foundation models for code.
\newblock \emph{arXiv preprint arXiv:2308.12950}, 2023.

\bibitem[Sakaguchi et~al.(2021)Sakaguchi, Bras, Bhagavatula, and Choi]{sakaguchi2021winogrande}
Sakaguchi, K., Bras, R.~L., Bhagavatula, C., and Choi, Y.
\newblock Winogrande: An adversarial winograd schema challenge at scale.
\newblock \emph{Communications of the ACM}, 64\penalty0 (9):\penalty0 99--106, 2021.

\bibitem[Schulhoff et~al.(2024)Schulhoff, Ilie, Balepur, Kahadze, Liu, Si, Li, Gupta, Han, Schulhoff, et~al.]{schulhoff2024prompt}
Schulhoff, S., Ilie, M., Balepur, N., Kahadze, K., Liu, A., Si, C., Li, Y., Gupta, A., Han, H., Schulhoff, S., et~al.
\newblock The prompt report: A systematic survey of prompting techniques.
\newblock \emph{arXiv preprint arXiv:2406.06608}, 2024.

\bibitem[Shen et~al.(2023)Shen, Jin, Huang, Liu, Dong, Guo, Wu, Liu, and Xiong]{shen2023large}
Shen, T., Jin, R., Huang, Y., Liu, C., Dong, W., Guo, Z., Wu, X., Liu, Y., and Xiong, D.
\newblock Large language model alignment: A survey.
\newblock \emph{arXiv preprint arXiv:2309.15025}, 2023.

\bibitem[Sorensen et~al.(2017)Sorensen, Elliott, Dixon, McDonald, nithum, and Cukierski]{adams2017toxic}
Sorensen, J., Elliott, J., Dixon, L., McDonald, M., nithum, and Cukierski, W.
\newblock Toxic comment classification challenge, 2017.
\newblock URL \url{https://kaggle.com/competitions/jigsaw-toxic-comment-classification-challenge}.

\bibitem[Sun et~al.(2023)Sun, Zhang, He, Li, Cheng, Yan, Liu, Shao, Tang, Zhao, et~al.]{sun2023moss}
Sun, T., Zhang, X., He, Z., Li, P., Cheng, Q., Yan, H., Liu, X., Shao, Y., Tang, Q., Zhao, X., et~al.
\newblock Moss: Training conversational language models from synthetic data.
\newblock \emph{arXiv preprint arXiv:2307.15020}, 7, 2023.

\bibitem[Taori et~al.(2023)Taori, Gulrajani, Zhang, Dubois, Li, Guestrin, Liang, and Hashimoto]{taori2023stanford}
Taori, R., Gulrajani, I., Zhang, T., Dubois, Y., Li, X., Guestrin, C., Liang, P., and Hashimoto, T.~B.
\newblock Stanford alpaca: An instruction-following llama model, 2023.

\bibitem[Team(2024)]{gemma_2024}
Team, G.
\newblock Gemma.
\newblock 2024.
\newblock \doi{10.34740/KAGGLE/M/3301}.
\newblock URL \url{https://www.kaggle.com/m/3301}.

\bibitem[Team et~al.(2024)Team, Mesnard, Hardin, Dadashi, Bhupatiraju, Pathak, Sifre, Rivi{\`e}re, Kale, Love, et~al.]{team2024gemma}
Team, G., Mesnard, T., Hardin, C., Dadashi, R., Bhupatiraju, S., Pathak, S., Sifre, L., Rivi{\`e}re, M., Kale, M.~S., Love, J., et~al.
\newblock Gemma: Open models based on gemini research and technology.
\newblock \emph{arXiv preprint arXiv:2403.08295}, 2024.

\bibitem[Touvron et~al.(2023)Touvron, Martin, Stone, Albert, Almahairi, Babaei, Bashlykov, Batra, Bhargava, Bhosale, et~al.]{touvron2023llama}
Touvron, H., Martin, L., Stone, K., Albert, P., Almahairi, A., Babaei, Y., Bashlykov, N., Batra, S., Bhargava, P., Bhosale, S., et~al.
\newblock Llama 2: Open foundation and fine-tuned chat models.
\newblock \emph{arXiv preprint arXiv:2307.09288}, 2023.

\bibitem[Tunstall et~al.(2023)Tunstall, Beeching, Lambert, Rajani, Rasul, Belkada, Huang, von Werra, Fourrier, Habib, Sarrazin, Sanseviero, Rush, and Wolf]{tunstall2023zephyr}
Tunstall, L., Beeching, E., Lambert, N., Rajani, N., Rasul, K., Belkada, Y., Huang, S., von Werra, L., Fourrier, C., Habib, N., Sarrazin, N., Sanseviero, O., Rush, A.~M., and Wolf, T.
\newblock Zephyr: Direct distillation of lm alignment, 2023.

\bibitem[Wan et~al.(2024)Wan, Bi, He, Zhang, Zhang, Sui, Xu, Jin, and Yu]{wan2024deep}
Wan, Y., Bi, Z., He, Y., Zhang, J., Zhang, H., Sui, Y., Xu, G., Jin, H., and Yu, P.
\newblock Deep learning for code intelligence: Survey, benchmark and toolkit.
\newblock \emph{ACM Computing Surveys}, 2024.

\bibitem[Wei et~al.(2022)Wei, Wang, Schuurmans, Bosma, Xia, Chi, Le, Zhou, et~al.]{wei2022chain}
Wei, J., Wang, X., Schuurmans, D., Bosma, M., Xia, F., Chi, E., Le, Q.~V., Zhou, D., et~al.
\newblock Chain-of-thought prompting elicits reasoning in large language models.
\newblock \emph{Advances in neural information processing systems}, 35:\penalty0 24824--24837, 2022.

\bibitem[Wei et~al.(2024)Wei, Yang, Song, Lu, Hu, Tran, Peng, Liu, Huang, Du, et~al.]{wei2024long}
Wei, J., Yang, C., Song, X., Lu, Y., Hu, N., Tran, D., Peng, D., Liu, R., Huang, D., Du, C., et~al.
\newblock Long-form factuality in large language models.
\newblock \emph{arXiv preprint arXiv:2403.18802}, 2024.

\bibitem[Wikipedia(2024)]{wikipedia}
Wikipedia.
\newblock Wikipedia, the free encyclopedia, 2024.
\newblock URL \url{https://www.wikipedia.org/}.
\newblock Accessed: 2024-05-24.

\bibitem[Ye et~al.(2023)Ye, Axmed, Pryzant, and Khani]{ye2023prompt}
Ye, Q., Axmed, M., Pryzant, R., and Khani, F.
\newblock Prompt engineering a prompt engineer.
\newblock \emph{arXiv preprint arXiv:2311.05661}, 2023.

\bibitem[Young et~al.(2024)Young, Chen, Li, Huang, Zhang, Zhang, Li, Zhu, Chen, Chang, et~al.]{young2024yi}
Young, A., Chen, B., Li, C., Huang, C., Zhang, G., Zhang, G., Li, H., Zhu, J., Chen, J., Chang, J., et~al.
\newblock Yi: Open foundation models by 01. ai.
\newblock \emph{arXiv preprint arXiv:2403.04652}, 2024.

\bibitem[Zhou et~al.(2022)Zhou, Muresanu, Han, Paster, Pitis, Chan, and Ba]{zhou2022large}
Zhou, Y., Muresanu, A.~I., Han, Z., Paster, K., Pitis, S., Chan, H., and Ba, J.
\newblock Large language models are human-level prompt engineers.
\newblock \emph{arXiv preprint arXiv:2211.01910}, 2022.

\end{thebibliography}
\bibliographystyle{icml2024}

\newpage
\appendix
\onecolumn

\section{Data Availability and Reproducibility}

The Open Artificial Knowledge (OAK) dataset is publicly available for research and development purposes at:

\textbf{Website}: \url{http://oakdataset.org/}

\textbf{Hugging Face}: \url{https://huggingface.co/datasets/tabularisai/oak}

The code for generating and managing the OAK dataset is available on GitHub, ensuring transparency and reproducibility:

\textbf{GitHub Repository}: \url{https://github.com/tabularis-ai/oak-dataset}



\section{LLM-based Prompt Creation}

\begin{table}[h!]
\scriptsize
\centering
\caption{The prompt creation using open-source LLMs and quality metrics.}
\label{tab:prompt_creation_1}
\begin{tabular}{
>{\centering\arraybackslash}p{2cm} 
>{\centering\arraybackslash}p{2cm} 
p{1.3cm} 
p{1.3cm} 
p{1.3cm} 
p{1.2cm} 
p{5.2cm}}
\toprule
\textbf{Main Category} & \textbf{Subcategory} & \multicolumn{3}{c}{\textbf{Constraints}} & \textbf{Model} & \textbf{Resulting Prompt} \\
\cmidrule(r){3-5}
 &  & \textbf{Quality} & \textbf{Length} & \textbf{Style} &  & \\
\toprule
European Government Stubs & Baltic States Political Systems & Impressive & Extensive & Advisory & Llama3 70B & \textit{\texttt{What are the key differences in the parliamentary systems of Estonia, Latvia, and Lithuania, and how have they evolved since the early 1990s, including their membership in the European Union, to shape the current political landscape of the Baltic states?}} \\
\midrule
Pantheism & Pantheism in Romantic Literature & Exemplary & Abbreviated & Argumentative & Mixtal 8B & \textit{\texttt{How did the transcendental and mystical elements of Pantheism influentially permeate the works of Romantic authors, such as Wordsworth, Coleridge, and Shelley, and what arguments can be made for its significance in shaping the literary movement?}} \\
\midrule
Lists of Tourist Attractions & Historic Landmarks & Exceptional & Extensive & Expressionistic & Llama3-8B & \textit{\texttt{Imagine yourself standing in the heart of a historic city, surrounded by centuries-old landmarks that whisper tales of the past. Create an exhaustive list of the world's most revered historic landmarks, including but not limited to, the Eiffel Tower, the Colosseum, and the Taj Mahal, highlighting their architectural significance, cultural importance, and the emotional resonance they evoke in tourists and locals alike.}} \\
\midrule
Sport and Apartheid in South Africa & The Role of Nelson Mandela in Sports and Apartheid & First-class & Extended & Factual & Mixtal 7x8b & \textit{\texttt{Discuss the significant impact of Nelson Mandela on the world of sports within the context of apartheid in South Africa. Explore his role in challenging this system of racial segregation and inequality, and how his influence extended to the realm of sports. Delve into the depth and breadth of his contributions, providing a comprehensive and informative analysis.}} \\
\midrule
Mathematical science occupations & Financial Mathematics & High & Summary-length & Urban & Gemma-7B & \textit{\texttt{Dive into the exciting realm of financial mathematics, exploring its applications in risk management, portfolio optimization, and quantitative trading. Highlight the diverse career paths and competitive salaries associated with this lucrative field.}}\\
\bottomrule
\end{tabular}
\end{table}

\begin{table}[h!]
\scriptsize
\centering
\caption{The prompt creation using open-source LLMs and quality metrics.}
\label{tab:prompt_creation_2}
\begin{tabular}{
>{\centering\arraybackslash}p{2cm} 
>{\centering\arraybackslash}p{2cm} 
p{1.3cm} 
p{1.3cm} 
p{1.3cm} 
p{1.2cm} 
p{5.2cm}}
\toprule
\textbf{Main Category} & \textbf{Subcategory} & \multicolumn{3}{c}{\textbf{Constraints}} & \textbf{Model} & \textbf{Resulting Prompt} \\
\cmidrule(r){3-5}
 &  & \textbf{Quality} & \textbf{Length} & \textbf{Style} &  & \\
\toprule
Days & Seasonal Holidays & Poor & Truncated & Realistic & Llama3 13B & \textit{\texttt{What are a few quickly mentioned seasonal holidays that one can observe during the year?}} \\
\midrule
Fiction About Television & Fictional TV Pilots & Unremarkable & Summary-length & Humorous & Mixtal 7x8B & \textit{\texttt{Write a ridiculously cheesy TV pilot about a group of superhero librarians who must band together to save the world from an evil organization of overdue-book wielding villains, in 100-150 words, with a dash of humor and wit.}} \\
\midrule
Futurologists & Artificial Intelligence & Poor & Short & Storytelling & Gemma-7B & \textit{\texttt{Imagine a world where artificial intelligence surpasses human intelligence. A renowned futurologist, standing at the forefront of AI research, gazes out upon the cityscape, pondering the implications of creating sentient machines. Ask them about the potential consequences of this revolution.}} \\
\midrule
Hausa history & Bagauda Dynasty & Low & Minimal& Wry& LlamA3-8b &\textit{\texttt{So, give me the CliffsNotes on Bagauda's reign in the Hausa history, yeah?}} \\
\midrule
Equipment & Household Appliances & Inferior & Below average length & Informal& Gemma 8B & \textit{\texttt{Write a brief, casual review of a new smart coffee maker for a friend who's moving into their first apartment, highlighting its ease of use and energy efficiency, in about 50-75 words}}\\
\midrule
Biology images & Botany & Mediocre & Bloated & Emotional & Llama3-70B & \textit{\texttt{With a touch of sentiment, describe an image deeply rooted in the world of biology, specifically focusing on the captivating field of botany. However, let's try to condense it a bit, shall we?}}\\
\midrule
Applied psychology & Organizational psychology & 	Weak	& Short	& Humorous & Gemma 8B & \textit{\texttt{Why does organizational psychology make work feel like a comedy show instead of a workplace? Discuss the hilarious quirks and awkward encounters that only organizational psychology can explain.}} \\
\midrule
Time in North Korea &	Traditional farming practices and seasons &	Very low &	Lengthy &	Informal & Mixtral 8x7B &
\textit{\texttt{Hey there, how much do you know about North Korea, specifically when it comes to their traditional farming practices and the seasons over there? Let's dive into this interesting topic. For instance, what kind of crops are typically grown in North Korea and how do the farming practices vary between seasons? And do these practices have any special significance in North Korean culture or society? Let's explore these questions and more.}} \\

\bottomrule
\end{tabular}
\end{table}

\begin{figure}[!h]
\begin{lstlisting}
# List of interest areas
interest_areas = ["technology", "history", "art", "science", ...]

def generate_prompt(categories, page_counts):
    interest = choose_random(categories, page_counts)
    
    if random_chance(0.09):
        random_title = fetch_random_wikipedia_title()
    else:
        page_titles = search_wikipedia(interest, 5)
        random_title = choose_random(page_titles)
    
    summary = fetch_wikipedia_summary(random_title, 2)
    analysis_type = choose_random([
        "summarize", "provide an in-depth analysis of", "contrast with another topic in the same field",
        "forecast future directions for", "explain in simple terms to a beginner", "examine the significance of",
        "describe through the lens of a specific philosophical perspective", "uncover the historical evolution of",
        "discuss the benefits and drawbacks of", "demonstrate with real-world cases"
    ])
    answer_length = choose_random([
        "a concise", "a brief", "a succinct", "a compact", "a short", "a detailed", "a comprehensive", 
        "an in-depth", "a thorough", "an extensive"
    ])
    
    prompt_templates = [
        f"Considering the vast array of knowledge and perspectives, please {analysis_type} the following topic: '{random_title}'. "
        f"Here's a brief introduction: {summary}. Your task is to provide {answer_length} answer, incorporating insights, examples, "
        f"or predictions that might illuminate the subject further for a diverse audience.",
        ...
        f"Amidst the tapestry of human knowledge, we invite you to {analysis_type} the captivating topic of '{random_title}'. "
        f"As a starting point, ponder this brief overview: {summary}. Your challenge is to create {answer_length} exploration "
        f"that illuminates the subject's depths, drawing upon a rich palette of perspectives, predictions, and real-world illustrations."
    ]
    
    return choose_random(prompt_templates)
\end{lstlisting}
\caption{Pseudocode for the dynamic prompt engineering algorithm using code: This algorithm generates diverse and contextually rich prompts for the OAK dataset by leveraging Wikipedia topics, randomized analysis types, and varied response lengths. It combines elements of randomization, topic selection, and template-based prompt construction to create a wide range of prompts suitable for synthetic data generation.}
\label{fig:programming_prompt}
\end{figure}

\section{Subtopic Expansion}
\label{app:subtopic}

The subtopic expansion process is a crucial step in the generation of the OAK dataset, aimed at enriching high-level topics with more detailed and diverse subtopics. This step leverages advanced language models, such as OpenAI’s GPT-4o, to ensure comprehensive coverage and quality. The Subtopic Expansion step is vital in creating a detailed, diverse, and high-quality synthetic dataset, forming the foundation for the subsequent steps in the OAK dataset generation pipeline.

The methodology for Subtopic Expansion involves several key steps: Initially, high-level topics are extracted from extensive human knowledge databases such as Wikipedia. This ensures a broad and diverse range of starting points for subtopic generation.

\begin{figure*}[ht]
    \centering
    \includegraphics[trim=8.0cm 9.2cm 8.0cm 9.0cm, clip, width=\textwidth]{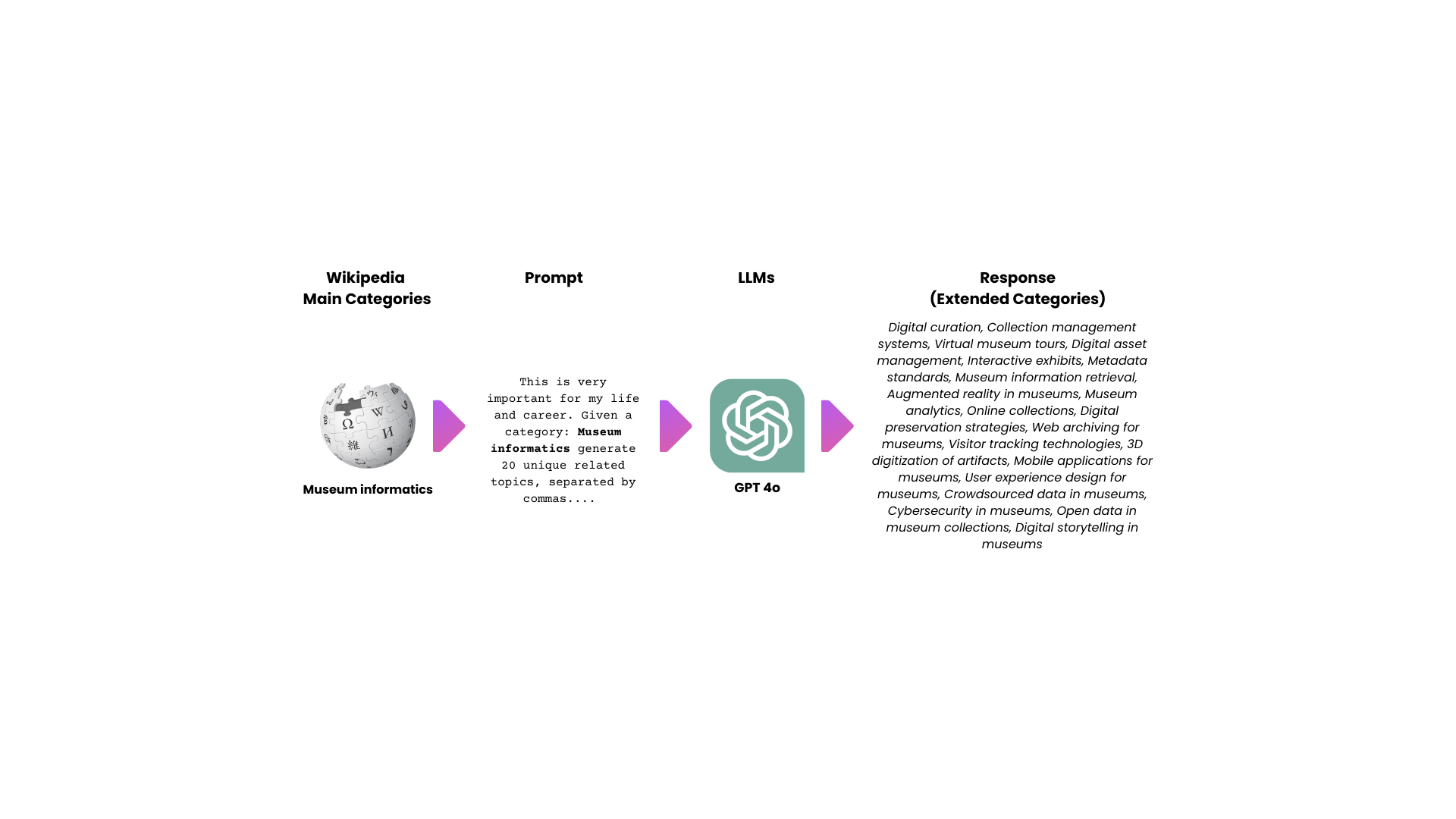}
    \caption{Overview of the extended topic generation step (Section \ref{sec:oak_dataset}). We prompt a GPT-4o model with a random main category from Wikipedia. The full prompt: \texttt{This is very important for my life and career. Given a category: \{category\}, generate 20 unique related topics, separated by commas. Do not include anything else, just the topics separated by commas, example: Topic1, Topic2, Topic3... Please adhere strictly to this format without numbering the topics.} The response is saved and later used for prompting and generation of the OAK dataset.}
    \label{fig:extended}
\end{figure*}

Using the extracted high-level topics, advanced language models like GPT-4o expand these topics into detailed subtopics. This method addresses both the diversity (\textbf{C1}) and quality (\textbf{C2}) challenges by mimicking real-world data variability and depth. An example of this step is presented in Figure \ref{fig:extended}. The prompt used for subtopic generation is presented in Figure \ref{fig:subtopic_prompt}.

\begin{figure}[h]
\begin{tcolorbox}[colback=gray!10!white, colframe=white!60!black, width=\textwidth, arc=3mm, auto outer arc, boxrule=0.75mm, title=Subtopic Expansion Prompt]

This is very important for my life and career. Given a category: \{category\}, generate 20 unique related topics, separated by commas. Do not include anything else, just the topics separated by commas, example: Topic1, Topic2, Topic3... Please adhere strictly to this format without numbering the topics.

\end{tcolorbox}
\caption{Subtopic Expansion Prompt: Prompt used to generate detailed subtopics from high-level topics.}
\label{fig:subtopic_prompt}
\end{figure}

\begin{figure}[h]
    \centering
    \begin{tcolorbox}[colback=gray!10!white, colframe=white!60!black, width=\textwidth, arc=3mm, auto outer arc, boxrule=0.75mm, title=Meta Prompt for OAK Dataset]
        \textbf{Your Task:} \\
        This is important for my career and life that you are an advanced prompt generator designed to create detailed, high-quality responses tailored to specific topics. For each input, you will generate a prompt that adheres strictly to predefined quality, length, and style criteria. \\
        
        \textbf{Scoring Criteria for Each Prompt:}
        \begin{itemize}
            \item \textbf{Quality}: Ranges from "low" to "superb", assessing the depth, relevance, and informativeness of the prompt.
            \item \textbf{Length}: Can be "brief", "concise", or "extensive", indicating the desired length of the prompt and the expected response.
            \item \textbf{Style}: Can be "informal", "formal", or "analytical", specifying the tone and format appropriate for the target audience.
        \end{itemize}
        
        \textbf{Example Prompts:}
        \begin{enumerate}
            \item \textbf{Quality: superb, Length: concise, Style: analytical} \\
            Input: [Main Topic: Technology, Subtopic: Artificial Intelligence in Healthcare] \\
            Output: ["Explore the profound impact of artificial intelligence on healthcare, focusing on its advancements in diagnostics, personalized medicine, and predictive analytics. Analyze both current practices and future prospects."]
            
            \item \textbf{Quality: basic, Length: brief, Style: informal} \\
            Input: [Main Topic: Environment, Subtopic: Renewable Energy Sources] \\
            Output: ["What are renewable energy sources and how do they benefit the environment?"]
        \end{enumerate}
        
        \textbf{Your Current Task:} \\
        This is important for my career and life. Given the following criteria and input, generate a new prompt:
        \begin{itemize}
            \item Quality: \{score\}
            \item Length: \{length\}
            \item Style: \{style\}
            \item Input: [Main Topic: \{category\}, Subtopic: \{subtopic\}]
        \end{itemize}
        Generate output in square brackets only.
    \end{tcolorbox}
    \caption{Meta Prompt: This prompt guides the creation of detailed and high-quality responses tailored to specific topics, following predefined criteria for quality, length, and style.}
    \label{fig:meta_prompt}
\end{figure}

\section{Generated Samples}

\begin{figure}
    \centering
\begin{shadowbox}[colback=gray!10]{Prompt}
Craft imaginative and insightful narratives describing the diverse regional cuisines across continents. Inject humor and cultural vibrancy into your descriptions, without sacrificing accuracy. Ensure each narrative captures the essence of the regional dishes and their significance within their respective cuisines.
\end{shadowbox}

\begin{shadowbox}{Response}
**Asian Fusion**

In the bustling streets of Bangkok, the aromas of sizzling street food wafted through the crowded alleys, teasing the taste buds of passersby. Amidst the cacophony of sounds and colors, one dish stood out: Pad Thai, a harmonious union of soft noodles, succulent shrimp, and crunchy peanuts. This iconic dish embodies the versatility and boldness of Thai cuisine, a delicate balance of sweet, sour, and spicy flavors.

**Latin American Fiesta**

In vibrant Rio de Janeiro, the rhythms of Samba met in harmony with the harmony of flavors in Feijoada, a hearty stew of black beans and pork. As the rich flavors simmered together, the vibrant atmosphere of Rio's beaches danced in perfect sync. The energy of Brazil's Afro-Latin fusion reverberated through every bite, transporting all who savored it to the sun-kissed shores.

...

**European Heritage**

In cozy Paris, the essence of French patisserie wafted from flaky croissants to delicate éclairs, each pastry a testament to centuries of artisanal craftsmanship. Butter, sugar, and precision dance in harmony, as French patisserie masters orchestrate an edible symphony of sweet and sour notes.

...

**African Fusion**

In Cape Town, South Africa, the flavors of Cape Malay cuisine blended in harmony with the rhythm of jazz and the colors of vibrant markets. Sampandrapé, a dish of tender lamb, potatoes, and spices, echoed the rich cultural tapestry of the Rainbow Nation. Flavors of coriander, cumin, and turmeric danced on the palate, echoing the cultural fusion of Portuguese, Dutch, and African traditions.

...

**Continental Fusion**

As the world's cuisines collided in the global kitchen, innovation and creativity merged the traditional with modern visions. In the eclectic streets of Tokyo, ramen, sushi, and curry rice fusioned, reflecting Japan's perpetual dialogue with the world. In the Mediterranean, from Greece to Turkey and beyond, the essence of olives, feta, and cucumbers reigned, a testament to the region's ancient connections.

...
\end{shadowbox}
    \caption{A random sample from the OAK dataset, generated using the Llama-8b model. The ellipsis (\textbf{...}) denotes text that has been omitted for brevity.}
    \label{fig:sample_1}
\end{figure}

\begin{figure}
    \centering
\begin{shadowbox}[colback=gray!10]{Prompt}
Explore the intricate representations of nature in Buddhist art, delving into the symbolism of trees, mountains, and rivers, and how they reflect the harmony between humanity and the natural world, while considering the role of environmentalism in modern Buddhist practices.
\end{shadowbox}

\begin{shadowbox}{Response}
**Intricate Representations of Nature in Buddhist Art**

Buddhist art has long celebrated the natural world, weaving intricate representations of trees, mountains, and rivers into its tapestry. These depictions transcend mere physical representation, conveying profound philosophical and spiritual insights.

**1. Trees:**

* Symbolize enlightenment, offering shelter and sustenance to travelers on the path to nirvana.
* Represent longevity, resilience, and interconnectedness.
* Often depicted with Banyan and Bodhi trees, significant to Buddhist tradition.

**2. Mountains:**

* Represent stability, permanence, and the overcoming of obstacles.
* Symbolize spiritual aspiration, leading towards the heavens.
* Depicted with rugged peaks, valleys, and waterfalls, reflecting their awe-inspiring grandeur.

**3. Rivers:**

* Represent the flow of life and the conscious momentum of the universe.
* Symbolize purification and rebirth, washing away impurities and guiding towards enlightenment.
* Often depicted with flowing water, reflecting the interconnectedness of all things.

**Harmony and Human-Nature Relationship:**

These natural elements highlight the harmonious relationship between humanity and the natural world. They represent a reciprocal interdependence, where humans are part of a larger ecological system. This is reflected in:

* **Buddhist ethics:** Emphasis on non-violence and respect for all life forms.
* **Zen practice:** Mindfulness and awareness cultivate a deep connection to the natural world.
* **Environmental conservation:** Buddhist teachings advocate for ecological responsibility and respect for the natural environment.

...

**Conclusion:**

The representation of nature in Buddhist art reflects a profound philosophical perspective on the harmony between humanity and the natural world. Environmentalism finds resonance within these traditions, urging a conscious responsibility for the care and preservation of our planet.
...
\end{shadowbox}
    \caption{A random sample from the OAK dataset, generated using the Gemma-7B model. The prompt is generated using the Llama3-8B model. The ellipsis (\textbf{...}) denotes text that has been omitted for brevity.}
    \label{fig:sample_2}
\end{figure}

\begin{figure}
    \centering
\begin{shadowbox}[colback=gray!10]{Prompt}
Design a thrilling skydiving experience for beginners, incorporating safety features and breathtaking views, in 50 words or less.
\end{shadowbox}

\begin{shadowbox}{Response}
Take the leap with our Thrillseeker Package! Jump from 10,000 feet with our experienced instructors, enjoying panoramic views of the coastline. Securely attached to the harness, you'll experience the rush of free-falling before opening your parachute and gliding back to solid ground. Safety first, thrills guaranteed!
\end{shadowbox}
    \caption{A random sample from the OAK dataset, generated using the Llama-70b model. The prompt is generated using the Mixtral 7x8B model.}
    \label{fig:sample_3}
\end{figure}

\begin{figure}
    \centering
    \includegraphics[width=\textwidth]{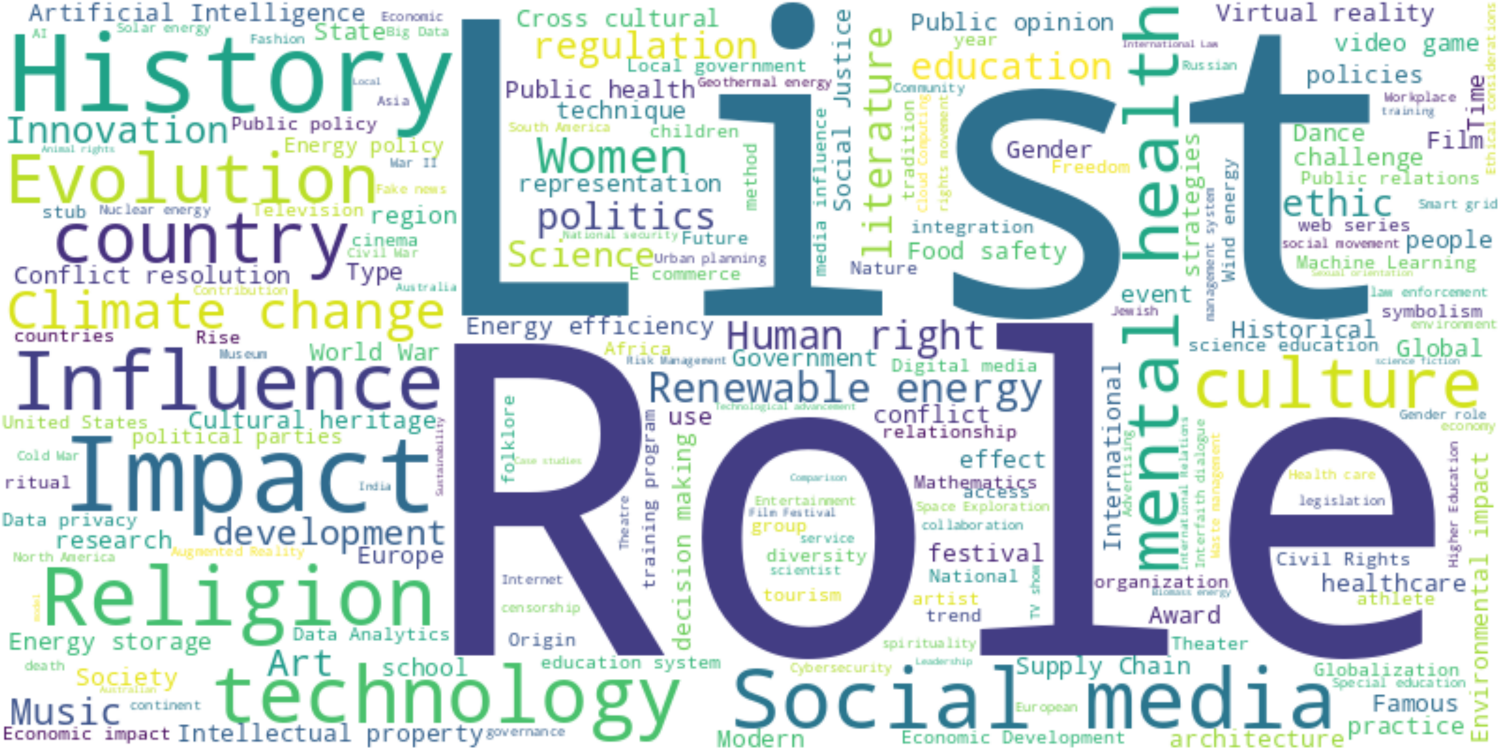}
    \caption{This word cloud visually represents the frequency of different subtopic generated in the Step $2$ (Section \ref{sec:oak_dataset}). For this step we utilize the GPT-4o model. Larger words indicate more frequently occurring topics, providing a quick overview of the most common themes. }
    \label{fig:enter-label}
\end{figure}

\newpage

\newpage
\newpage
\newpage

\end{document}